\documentclass[runningheads]{llncs}
\usepackage[T1]{fontenc}
\usepackage{graphicx}
\usepackage{xcolor}
\usepackage{amsmath}
\usepackage{amssymb}
\usepackage{booktabs}

\usepackage{multirow}

\definecolor{wacvblue}{rgb}{0.21,0.49,0.74}
\definecolor{green}{RGB}{34, 139 , 34}
\newcommand{\dgssm}{\textsc{DGSSM}}
\newcommand{\modelname}{\textsc{DGSSM }}

\begin{document}
\title{\dgssm: Diffusion guided state-space models for multimodal salient object detection}
%
%
\author{Suklav Ghosh\inst{1}\orcidID{0009-0002-7370-9152} \and
Arijit Sur\inst{1}\orcidID{0000-0002-9038-8138} \and
Pinaki Mitra\inst{1}\orcidID{0000-0002-8254-8234}}
\authorrunning{S. Ghosh et al.}
%
\institute{Dept. of Computer Science and Engineering, Indian Institute of Technology, Guwahati\\
\email{\{suklav,arijit,pinaki\}@iitg.ac.in}}
\maketitle              

\begin{abstract}

Salient object detection (SOD) requires modeling both long-range contextual dependencies and fine-grained structural details, which remains challenging for convolutional, transformer-based, and Mamba-based state space models. While recent Mamba-based state space approaches enable efficient global reasoning, they often struggle to recover precise object boundaries. In contrast, diffusion models capture strong structural priors through iterative denoising, but their use in discriminative dense prediction is still limited due to computational cost and integration challenges.
In this work, we propose \dgssm, a diffusion-guided state space (Mamba) framework that formulates multimodal salient object detection as a progressive denoising process. The framework integrates diffusion structural priors with multi-scale state space encoding, adaptive saliency prompting, and an iterative Mamba diffusion refinement mechanism to improve boundary accuracy. A boundary-aware refinement head and self-distillation strategy further enhance spatial coherence and feature consistency.
Extensive experiments on 13 public benchmarks across RGB, RGB-D, and RGB-T settings demonstrate that \dgssm\ consistently outperforms state-of-the-art methods across multiple evaluation metrics while maintaining a compact model size. These results suggest that diffusion-guided state space modeling is an effective and generalizable paradigm for multimodal dense prediction tasks.


\end{abstract}

\vspace{-12pt}
\section{Introduction}

Salient object detection (SOD) aims to identify and segment the most visually 
prominent regions in a scene~\cite{Englebert2022BRCAM,Lin2022Lightweight,Zhang2022CSMAN}. It provides essential cues for a wide range of vision applications such as object tracking~\cite{Zhou2021SaliencyTracking}, semantic segmentation~\cite{Wan2024SIGMA}, 
image enhancement~\cite{Miangoleh2023SaliencyEnhancement}, autofocus~\cite{Jiang2024LFTransformer}, and model evaluation~\cite{Jiang2024VLMAssessment}. 
Despite sustained progress over the past decade, SOD remains challenging in complex real-world scenarios involving cluttered backgrounds, low contrast, ambiguous object boundaries, and modality-specific noise.
\begin{figure}
    \centering
    \includegraphics[width=0.75\linewidth]{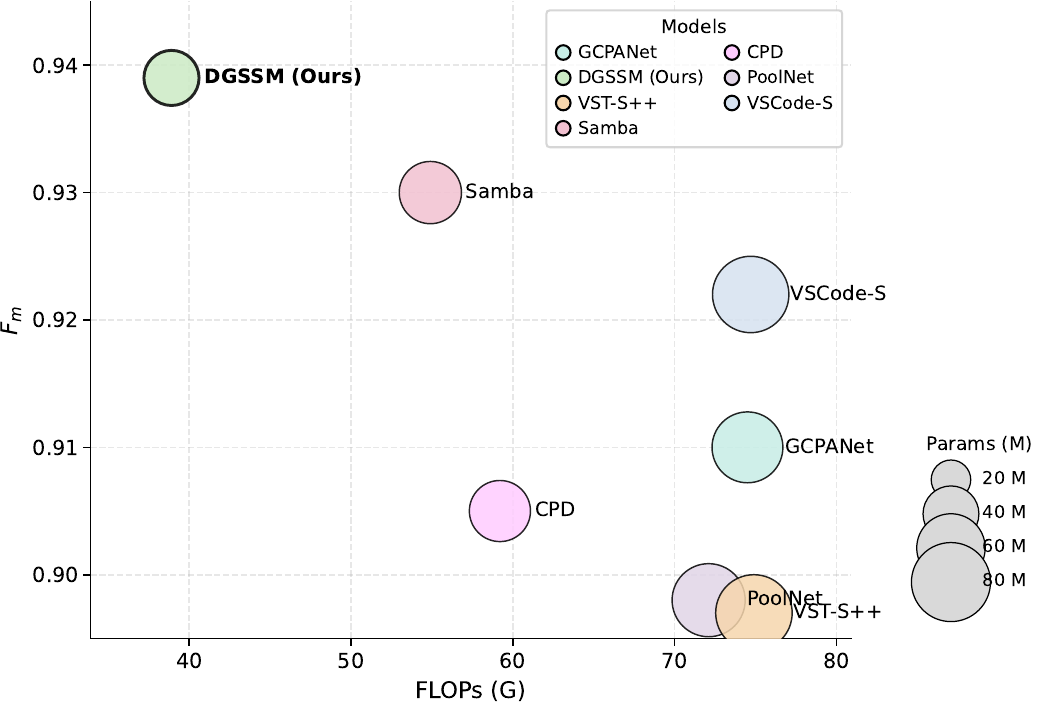}
    \caption{Accuracy efficiency trade-off between FLOPs, performance ($F_m$), and parameters. Bubble area denotes model size. DGSSM achieves superior accuracy with lower computational cost.}
\vspace{-13pt}
    \label{fig:intro}
\end{figure}

Recent advances in SOD have largely been driven by deep learning models based on 
convolutional neural networks (CNNs) and transformers. These approaches have achieved 
strong performance across diverse settings, including RGB, RGB-D, and RGB-T SOD 
tasks~\cite{Li2024Depth,Zhang2023SaliencyPrototype,Luo2024VSCode,He2025Samba}, as well as video-based extensions such as VSOD and 
RGB-D VSOD~\cite{Luo2024VSCode,He2025Samba}. CNN-based architectures are computationally 
efficient and scalable, benefiting from linear complexity. However, their limited 
receptive fields restrict the ability to capture long-range dependencies, often 
leading to fragmented saliency maps or incomplete object representations. Transformer
based models address this limitation by leveraging global self-attention, enabling 
richer contextual reasoning. Nevertheless, the quadratic complexity of self-attention 
introduces significant computational and memory overhead, which is particularly 
problematic for high-resolution dense prediction. Although efficient transformer 
variants, such as swin transformer~\cite{Liu2021Swin} and MobileViT~\cite{Mehta2022MobileViT}, have been proposed, they frequently sacrifice global modeling capacity to achieve efficiency,  resulting in an imperfect trade-off.

More recently, state space models (SSMs) have emerged as a promising alternative for 
efficient global modeling. In particular, Mamba~\cite{Gu2023Mamba} introduces a selective scanning mechanism that enables long-range dependency modeling with linear complexity and favorable hardware efficiency. Building upon this foundation, several visual adaptations of Mamba~\cite{Liu2024VMamba,Zhu2024VisionMamba} and task-specific variants~\cite{Dong2025FusionMamba,Wan2024SIGMA,Zhou2024DMM}  have demonstrated competitive performance across a variety of vision tasks. Thus, it can be concluded that SSMs provide a compelling backbone for dense prediction problems that require both efficiency and global context. However, despite their potential, existing Mamba-based models remain underexplored in salient object detection, where precise boundary localization and structural consistency are critical.
In parallel, diffusion models have gained considerable attention due to their strong ability to capture structural priors through iterative denoising~\cite{Mei2024CoDi,Zhang2023ConditionalControl}. Diffusion processes have been widely studied for image generation and, more recently, for segmentation-oriented tasks. Whereas, their application to discriminative dense prediction problems such as SOD remains limited. A key challenge lies in integrating diffusion models into efficient feature extraction pipelines without incurring prohibitive computational cost~\cite{Mei2024CoDi}. Importantly, the progressive refinement property of diffusion models is well suited to addressing one of the core challenges in SOD accurate and consistent boundary prediction, yet this potential has not been fully exploited in existing approaches.

Motivated by these observations, we argue that salient object detection can be more 
effectively formulated as a progressive denoising process that combines deterministic 
global modeling with stochastic structural refinement. To this end, we propose 
\dgssm, a diffusion-guided state space framework that unifies the efficiency of 
Mamba-based sequential modeling with the structural expressiveness of diffusion 
priors. By introducing diffusion-derived guidance into multi-scale state space 
encoding and refining predictions through iterative denoising, \dgssm\ achieves a 
favorable balance between global context modeling, boundary precision, and 
computational efficiency w.r.t. state-of-the-art methods (Fig. \ref{fig:intro}).
In summary, this work makes the following contributions:
\begin{itemize}
    \item We propose \dgssm, a novel diffusion-guided state space framework that 
    reformulates salient object detection as a progressive denoising process. It  
    effectively bridges deterministic state space modeling and stochastic 
    diffusion-based refinement.
    \item We introduce a diffusion structural prior and adaptive saliency prompting 
    to guide multi-scale selective scanning in Mamba-based encoders. It enhances global 
    coherence and boundary awareness.
    \item We design an iterative Mamba diffusion refinement mechanism and a 
    boundary-aware refinement head. These progressively improve saliency predictions, 
    particularly along object contours.
\end{itemize}

\section{Related Work}
\vspace{-8pt}

\subsection{Deep Learning Based Salient Object Detection}
\vspace{-8pt}

\subsubsection{Unimodal SOD:}
Early research on salient object detection primarily focused on the RGB modality and explored a variety of architectural strategies to enhance localization accuracy. Representative approaches include boundary-aware modeling~\cite{Zhao2019EGNet}, progressive feature refinement~\cite{Wu2022EDN}, and attention-based mechanisms designed to emphasize informative regions~\cite{Wang2023PRO}. With the advent of transformer architectures, several works~\cite{Liu2024VSTpp,Ma2023BRF,Zhuge2022Integrity} have leveraged global self-attention to model long-range dependencies, leading to notable performance improvements. Despite their success, these methods often encounter difficulties in challenging scenarios, such as complex backgrounds or low-contrast environments, where fine-grained structural cues and precise boundary delineation are critical.
\vspace{-13pt}
\subsubsection{Multimodal SOD:}
To address the limitations of RGB-only models in complex scenes, a growing body of research has incorporated auxiliary modalities, most notably depth~\cite{Chen2024RGBD,Fu2020JLDCF,Hu2024CMF,Lee2022SPSN,Sun2023CATNet,Zhou2021Specificity} and thermal imagery~\cite{Chen2022CGMDRNet,Cong2022Thermal,Huo2021ECSRNet,Tu2021MIDecoder,Zhang2023SaliencyPrototype}. These modalities provide complementary geometric or radiometric information that enhances robustness under adverse conditions. For instance, Fu et al.~\cite{Fu2020JLDCF} employ a siamese architecture to extract shared representations from RGB and depth inputs, improving saliency estimation in cluttered scenes. Tu et al.~\cite{Tu2021MIDecoder} propose a dual-decoder framework that explicitly models diverse interaction patterns between RGB and thermal modalities, yielding more discriminative feature fusion.
Beyond static imagery, video salient object detection (VSOD) introduces additional challenges due to dynamic motion patterns and temporal inconsistencies. To exploit temporal cues, several methods~\cite{Guo2024UNITR,Zhao2024MotionMemory} directly model frame-to-frame interactions to extract motion-aware representations. Alternatively, other approaches~\cite{Ji2021FullDuplex,Liu2023CSTTransformer} rely on optical flow estimation to capture complementary motion information between adjacent frames.
The integration of depth information into VSOD has further expanded this line of research. Recent studies~\cite{Lu2022DepthCoop} demonstrate that depth cues can significantly improve temporal saliency modeling, motivating the development of RGB-D VSOD. With the increasing availability of RGB-D video datasets~\cite{Lin2024VIDSOD100,Mou2024RGBDVideo}, methods such as DCTNet+~\cite{Mou2024RGBDVideo} and ATFNet~\cite{Lin2024VIDSOD100} have reported promising results, highlighting the effectiveness of depth-assisted temporal reasoning.
\vspace{-8pt}
\subsection{Diffusion Models}
Diffusion probabilistic models have recently achieved remarkable success across a wide range of vision tasks, including image generation~\cite{Mei2024CoDi,Zhang2023ConditionalControl}, semantic segmentation~\cite{Wan2024SIGMA}, object detection~\cite{Jiang2024LFTransformer,Moser2024DiffusionSurvey}, and super-resolution~\cite{Moser2024DiffusionSurvey}. These models are particularly effective at capturing complex data distributions through iterative denoising, enabling the generation of high-quality and structurally coherent outputs.
Inspired by these advances, recent studies have begun exploring diffusion models for challenging perception tasks where boundary prediction and structural consistency are critical. In environments characterized by severe visual degradation, such as underwater scenes, accurate boundary localization remains a longstanding challenge due to scattering, color distortion, and noise. The multi-step refinement process inherent to diffusion models naturally produces intermediate predictions that progressively approximate the desired output. Leveraging these intermediate representations can enhance object discrimination and boundary reliability. 

However, a key open problem lies in assessing and exploiting the reliability of intermediate diffusion predictions during the denoising trajectory, particularly in discriminative tasks.
\vspace{-8pt}
\subsection{Visual Mamba and State Space Models}
Motivated by the success of Mamba architectures in language modeling, Zhu et al.~\cite{Zhu2024VisionMamba} extend state space models to the vision domain by introducing Vision Mamba (ViM), which employs bidirectional state space formulations for efficient global context modeling. Building on this direction, Liu et al.~\cite{Liu2024VMamba} propose visual state space blocks and construct a hierarchical backbone, VMamba, which demonstrates strong performance across diverse vision tasks, including semantic segmentation~\cite{Wan2024SIGMA} and object detection~\cite{Dong2025FusionMamba,Zhou2024DMM}.

Subsequent works have further investigated the role of spatial scanning strategies in visual state space models. Zhao et al.~\cite{Zhao2024RSMamba} augment the conventional four-directional scanning mechanism with additional diagonal directions, thereby enlarging the effective receptive field and enhancing spatial feature aggregation. Yang et al.~\cite{Yang2024PlainMamba} propose PlainMamba, which adopts a continuous two dimensional scanning strategy to ensure spatial continuity in the generated sequences. These advances highlight the importance of scanning order and spatial coherence in state space vision models, motivating further exploration of guided and structure-aware scanning mechanisms.
\vspace{-10pt}
\section{Proposed Method}
\begin{figure}
    \centering
    \includegraphics[width=1.0\linewidth]{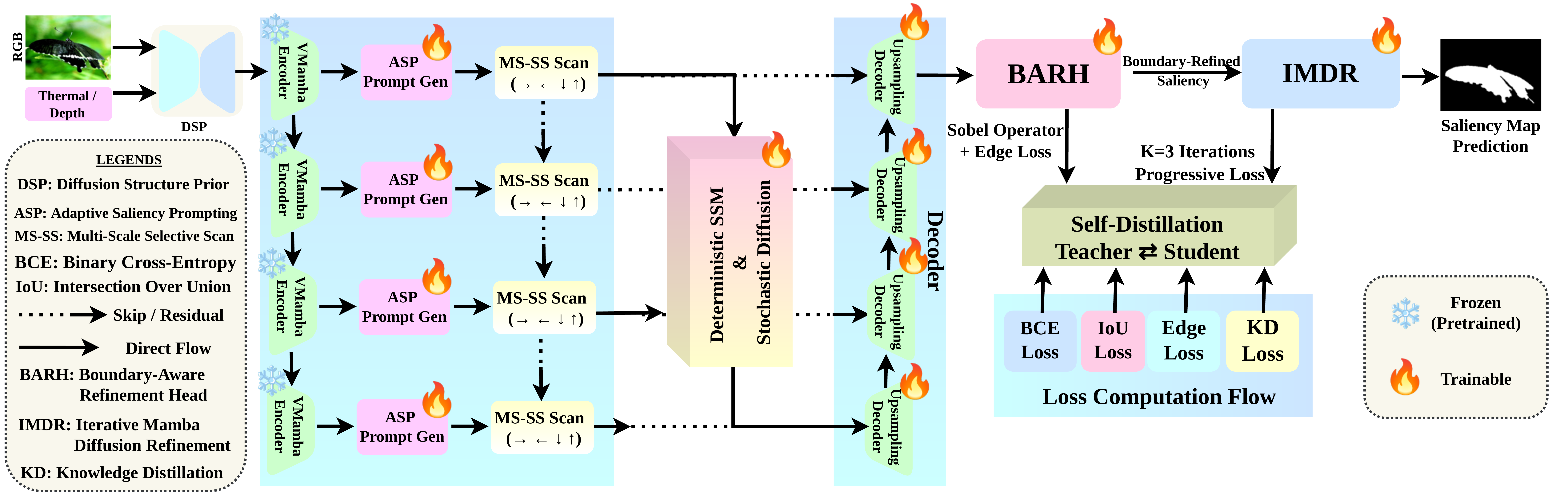}
    \vspace{-15pt}
    \caption{Overall architecture of \dgssm. A diffusion structural prior guides a hierarchical state space encoder with adaptive saliency prompting and multi-scale selective scanning to capture global context and structural cues. The decoder produces a coarse saliency map. It is further refined by a boundary-aware head and an iterative Mamba diffusion refinement module, yielding accurate and boundary-preserving saliency predictions.
}
\vspace{-25pt}
    \label{fig:architecture}
\end{figure}

\vspace{-8pt}
This section presents \dgssm, a diffusion-guided state space framework for 
salient object detection that tightly couples deterministic sequential modeling 
with stochastic structural refinement. As illustrated in Fig.~\ref{fig:architecture}, 
the proposed architecture integrates diffusion-derived priors, adaptive saliency 
prompting, multi-scale selective scanning, and iterative refinement within a unified 
encoder decoder design. Unlike conventional feed-forward saliency predictors, 
\modelname formulates saliency estimation as a progressive denoising process governed 
by both state space dynamics and diffusion-guided corrections.
\vspace{-10pt}
\subsection{Overview and Problem Definition}
\vspace{-8pt}
Given an input image $I \in \mathbb{R}^{H \times W \times C}$, optionally accompanied 
by an auxiliary modality $I^{a}$ (depth or thermal), our goal is to predict a dense 
saliency map $S \in [0,1]^{H \times W}$. Rather than modeling this task as a direct 
mapping, we interpret saliency inference as a trajectory. A trajectory that evolves from coarse, structurally uncertain representations toward a boundary-accurate prediction:
\begin{equation}
    S = \mathcal{G}\big( \mathcal{T}(I, I^{a}) \big),
\end{equation}
where $\mathcal{T}(\cdot)$ denotes diffusion-guided state space feature extraction 
and $\mathcal{G}(\cdot)$ represents iterative refinement and decoding.
\vspace{-8pt}
\subsection{Diffusion Structural Prior (DSP)}
\label{sec:dsp}
\vspace{-8pt}
To inject global structural awareness into the encoder, we introduce a 
Diffusion Structural Prior (DSP) module. A pretrained diffusion model is used 
to generate a compact latent representation $z_0$ of the input, which undergoes 
a forward noising process:
\begin{equation}
    z_t = \sqrt{\alpha_t} z_0 + \sqrt{1-\alpha_t} \epsilon, \quad 
    \epsilon \sim \mathcal{N}(0, \mathbf{I}),
\end{equation}
where $\alpha_t$ follows a predefined noise schedule. Instead of executing the full 
reverse process, we perform a limited number of denoising steps to obtain an 
intermediate latent $z_{t^\star}$ that preserves object layout and boundary cues.

This latent serves as a frozen structural prior and is injected into each encoder 
stage:
\begin{equation}
    \tilde{X}^{(l)} = \mathrm{Concat}\big(X^{(l)}, z_{t^\star}\big),
\end{equation}
where $X^{(l)}$ denotes the feature map at the $l$-th encoder stage.
\vspace{-12pt}
\subsection{Adaptive Saliency Prompting}
\label{sec:asp}
\vspace{-8pt}
At each encoder level, we introduce Adaptive Saliency Prompting (ASP) to dynamically 
modulate state space transitions. Given a feature map $\tilde{X}^{(l)}$, ASP produces 
a prompt vector $p^{(l)}$ via global context aggregation:
\begin{equation}
    p^{(l)} = \phi\!\left(\mathrm{GAP}\left(\tilde{X}^{(l)}\right)\right),
\end{equation}
where $\phi(\cdot)$ denotes a lightweight projection network. The prompt is used to 
condition the state space update parameters, enabling saliency-aware adaptation of 
the sequential dynamics.
\vspace{-8pt}
\subsection{Multi-Scale Selective State Space Scan}
\label{sec:msss}
\vspace{-8pt}
Each encoder stage employs a Multi-Scale Selective Scan (MS-SS) mechanism to model 
long-range spatial dependencies. For a given scanning direction $d \in 
\{\rightarrow, \leftarrow, \uparrow, \downarrow\}$, the state update follows:
\begin{equation}
    h_{i+1}^{(d)} = \mathbf{A}^{(l)} h_i^{(d)} + \mathbf{B}^{(l)} x_i,
\end{equation}
where $h_i^{(d)}$ is the hidden state at spatial index $i$ and $x_i$ denotes the input 
token. Scans are performed at multiple spatial resolutions, and the resulting 
representations are aggregated:
\begin{equation}
    X_{\text{MS}}^{(l)} = \sum_{s=1}^{S} \sum_{d} \mathcal{S}_{s,d}\big(X^{(l)}\big),
\end{equation}
where $\mathcal{S}_{s,d}$ denotes scanning at scale $s$ and direction $d$.

\subsection{Deterministic Stochastic Interaction}
\label{sec:dsbridge}

A central component of \modelname is the interaction between deterministic state-space 
modeling and stochastic diffusion guidance. Let $F_m^{(l)}$ denote the output of the 
MS-SS encoder at stage $l$, and $F_d$ the diffusion prior. Their interaction is 
modeled as:
\begin{equation}
    F^{(l)} = F_m^{(l)} + \lambda_l \cdot \Psi(F_d),
\end{equation}
where $\Psi(\cdot)$ projects the diffusion latent into the feature space and 
$\lambda_l$ is a learnable scale factor. This formulation allows diffusion-derived 
structure to bias deterministic transitions without overwhelming them.
\vspace{-8pt}
\subsection{Decoder and Boundary-Aware Refinement Head}
\label{sec:barh}

Multi-scale features are progressively upsampled by a state space decoder to 
produce a coarse saliency map $\hat{S}_0$. To explicitly enhance boundary accuracy, 
we introduce a Boundary-Aware Refinement Head (BARH). Edge cues $E$ are extracted 
using a Sobel operator:
\begin{equation}
    E = \|\nabla_x \hat{S}_0\| + \|\nabla_y \hat{S}_0\|.
\end{equation}
The edge features are fused with decoder outputs through a refinement module to 
obtain a boundary-enhanced prediction $\hat{S}_b$.

\subsection{Iterative Mamba Diffusion Refinement}
\label{sec:imdr}

The refined saliency is further processed through Iterative Mamba Diffusion 
Refinement (IMDR). Starting from $\hat{S}_b$, the model performs $K$ refinement 
iterations:
\begin{equation}
    \hat{S}_{k+1} = \hat{S}_{k} + \mathcal{R}\big(\hat{S}_{k}, F_d\big),
\end{equation}
where $\mathcal{R}(\cdot)$ denotes a lightweight diffusion-guided correction module. 
Progressive supervision is applied at each iteration to stabilize optimization.

\subsection{Self-Distillation and Training Objective}
\label{sec:loss}

To improve feature consistency across scales, we employ self-distillation by treating 
the deepest encoder stage as a teacher. For a student feature $F_s^{(l)}$ and teacher 
feature $F_t$, the distillation loss is defined as:
\begin{equation}
    \mathcal{L}_{\mathrm{KD}} = \sum_{l} \left\| \eta(F_s^{(l)}) - \eta(F_t) \right\|_2^2.
\end{equation}

The overall training objective combines saliency supervision, boundary constraints, 
diffusion consistency, and distillation:
\begin{equation}
\begin{aligned}
    \mathcal{L}_{\text{total}} =\;& 
    \mathcal{L}_{\text{BCE}} + \mathcal{L}_{\text{IoU}} 
    + \gamma \mathcal{L}_{\text{edge}} \\
    &+ \delta \mathcal{L}_{\text{KD}}
    + \sum_{k=1}^{K} \omega_k \mathcal{L}_{\text{prog}}^{(k)} ,
\end{aligned}
\end{equation}
where $\gamma$, $\delta$, and $\omega_k$ control the contribution of each term.

\begin{table*}[t]
\centering
\caption{Comparison of our DGSSM against other SOTA RGB SOD methods on five benchmark datasets. The best results are highlighted in \textcolor{red}{red}, \textcolor{blue}{blue}, and \textcolor{green}{green}, respectively.}
\vspace{-10pt}
\label{tab:rgb}
\resizebox{\textwidth}{!}{
\begin{tabular}{l|c|ccc|ccc|ccc|ccc|ccc}
\toprule
\multirow{2}{*}{\textbf{Method}} & \multirow{2}{*}{\textbf{Params (M)}} & \multicolumn{3}{c|}{\textbf{DUTS}~\cite{Wang2017ImageLevel}} & \multicolumn{3}{c|}{\textbf{DUT-O}~\cite{Yang2013GraphManifold}} & \multicolumn{3}{c|}{\textbf{HKU-IS}~\cite{Li2015MultiScale}} & \multicolumn{3}{c|}{\textbf{PASCAL-S}~\cite{Li2014Secrets}} & \multicolumn{3}{c}{\textbf{ECSSD}~\cite{Yan2013Hierarchical}} \\
 & & $S_m\uparrow$ & $F_m\uparrow$ & $E_m\uparrow$ & $S_m\uparrow$ & $F_m\uparrow$ & $E_m\uparrow$ & $S_m\uparrow$ & $F_m\uparrow$ & $E_m\uparrow$ & $S_m\uparrow$ & $F_m\uparrow$ & $E_m\uparrow$ & $S_m\uparrow$ & $F_m\uparrow$ & $E_m\uparrow$ \\
\midrule
\multicolumn{17}{c}{ \textit{\textbf{CNN-based}}} \\
\midrule

CSF-R2~\cite{Gao2020Efficient}    & 36.53  & 0.890 & 0.869 & 0.929 & 0.838 & 0.775 & 0.869 & -     & -     & -     & 0.863 & 0.839 & 0.885 & 0.931 & 0.942 & 0.960 \\
EDN~\cite{Wu2022EDN}       & 42.85  & 0.892 & 0.893 & 0.933 & 0.849 & 0.821 & 0.884 & 0.924 & 0.940 & 0.963 & 0.864 & 0.879 & 0.907 & 0.927 & 0.950 & 0.957 \\
ICON-R~\cite{Zhuge2022Integrity}    & 33.09  & 0.890 & 0.876 & 0.931 & 0.845 & 0.799 & 0.884 & 0.920 & 0.931 & 0.960 & 0.862 & 0.844 & 0.888 & 0.928 & 0.943 & 0.960 \\
MENet~\cite{Wang2023Multiple}     & 27.83  & 0.905 & 0.895 & 0.943 & 0.850 & 0.792 & 0.879 & 0.927 & 0.939 & 0.965 & 0.871 & 0.848 & 0.892 & 0.927 & 0.938 & 0.956 \\
\midrule
\multicolumn{17}{c}{\textit{\textbf{Transformer-based}}} \\
\midrule

ICON-S~\cite{Zhuge2022Integrity}    & 94.30  & 0.917 & 0.911 & 0.960 & 0.869 & 0.830 & 0.906 & 0.936 & 0.947 & 0.974 & 0.885 & 0.860 & 0.903 & 0.941 & 0.954 & 0.971 \\
BBRF~\cite{Ma2023BRF}      & 74.40  & 0.908 & 0.905 & 0.951 & 0.855 & 0.820 & 0.898 & 0.935 & 0.946 & 0.936 & 0.871 & 0.884 & 0.925 & 0.939 & 0.957 & 0.972 \\
VST-S++~\cite{Liu2024VSTpp}   & 74.90  & 0.909 & 0.897 & 0.947 & 0.859 & 0.813 & 0.890 & 0.932 & 0.941 & 0.969 & 0.880 & 0.859 & 0.901 & 0.939 & 0.951 & 0.969 \\
VSCode-S~\cite{Luo2024VSCode}  & 74.72  & \textcolor{green}{0.926} & \textcolor{green}{0.922} & \textcolor{green}{0.960} & \textcolor{green}{0.877} & \textcolor{green}{0.840} & \textcolor{green}{0.912} & \textcolor{green}{0.940} & \textcolor{green}{0.951} & \textcolor{green}{0.974} & \textcolor{green}{0.887} & \textcolor{green}{0.864} & \textcolor{green}{0.904} & \textcolor{green}{0.949} & \textcolor{green}{0.959} & \textcolor{green}{0.974} \\
Samba~\cite{He2025Samba} & 49.59 & \textcolor{blue}{0.932} & \textcolor{blue}{0.930} & \textcolor{blue}{0.966} & \textcolor{blue}{0.889} & \textcolor{blue}{0.859} & \textcolor{blue}{0.922} & \textcolor{blue}{0.945} & \textcolor{blue}{0.956} & \textcolor{blue}{0.978} & \textcolor{blue}{0.892} & \textcolor{blue}{0.896} & \textcolor{blue}{0.931} & \textcolor{blue}{0.953} & \textcolor{blue}{0.965} & \textcolor{blue}{0.978} \\
\midrule
\textbf{DGSSM (Ours)} & \textbf{38.92} & \textbf{\textcolor{red}{0.947}} & \textbf{\textcolor{red}{0.939}} & \textbf{\textcolor{red}{0.978}} & \textbf{\textcolor{red}{0.903}} & \textbf{\textcolor{red}{0.869}} & \textbf{\textcolor{red}{0.930}} & \textbf{\textcolor{red}{0.951}} & \textbf{\textcolor{red}{0.959}} & \textbf{\textcolor{red}{0.981}} & \textbf{\textcolor{red}{0.899}} & \textbf{\textcolor{red}{0.901}} & \textbf{\textcolor{red}{0.934}} & \textbf{\textcolor{red}{0.963}} & \textbf{\textcolor{red}{0.973}} & \textbf{\textcolor{red}{0.983}} \\
\bottomrule
\end{tabular}
}
\end{table*}

\begin{table*}[t]
\centering
\caption{Comparison of our DGSSM against other SOTA RGB-D SOD methods on five benchmark datasets. The best results are highlighted in \textcolor{red}{red}, \textcolor{blue}{blue}, and \textcolor{green}{green}, respectively.}
\vspace{-10pt}
\label{tab:rgbd}
\resizebox{\textwidth}{!}{
\begin{tabular}{l|c|ccc|ccc|ccc|ccc|ccc}
\toprule
\multirow{2}{*}{\textbf{Method}} & \multirow{2}{*}{\textbf{Params (M)}} & \multicolumn{3}{c|}{\textbf{NJUD}~\cite{Ju2014DepthSaliency}} & \multicolumn{3}{c|}{\textbf{NLPR}~\cite{Peng2014RGBD}} & \multicolumn{3}{c|}{\textbf{SIP}~\cite{Fan2020RGBDReview}} & \multicolumn{3}{c|}{\textbf{STERE}~\cite{Niu2012Stereopsis}} & \multicolumn{3}{c}{\textbf{DUTLF-D}~\cite{Piao2019DMSRAN}} \\
 & & $S_m\uparrow$ & $F_m\uparrow$ & $E_m\uparrow$ & $S_m\uparrow$ & $F_m\uparrow$ & $E_m\uparrow$ & $S_m\uparrow$ & $F_m\uparrow$ & $E_m\uparrow$ & $S_m\uparrow$ & $F_m\uparrow$ & $E_m\uparrow$ & $S_m\uparrow$ & $F_m\uparrow$ & $E_m\uparrow$ \\
\midrule
\multicolumn{16}{c}{\textit{\textbf{CNN-based}}} \\
\midrule

JL-DCF~\cite{Fu2020JLDCF}    & 143.52 & 0.877 & 0.892 & 0.941 & 0.931 & 0.918 & 0.965 & 0.885 & 0.894 & 0.931 & 0.900 & 0.895 & 0.942 & 0.894 & 0.891 & 0.927 \\
SP-Net~\cite{Zhou2021Specificity}    & 67.88  & 0.925 & 0.928 & 0.957 & 0.927 & 0.919 & 0.962 & 0.894 & 0.904 & 0.933 & 0.907 & 0.906 & 0.949 & 0.895 & 0.899 & 0.933 \\
SPSN~\cite{Lee2022SPSN}      & -      & 0.925 & 0.930 & 0.956 & 0.918 & 0.921 & 0.952 & 0.923 & 0.912 & \textcolor{green}{0.960} & 0.892 & 0.900 & 0.936 & 0.907 & 0.902 & 0.945 \\
\midrule
\multicolumn{16}{c}{\textit{\textbf{Transformer-based}}} \\
\midrule

SwinNet-B~\cite{Liu2021SwinNet} & 199.18 & 0.920 & 0.924 & 0.956 & \textcolor{green}{0.941} & \textcolor{green}{0.936} & \textcolor{green}{0.974} & 0.911 & 0.927 & 0.950 & 0.919 & 0.918 & 0.956 & 0.918 & 0.920 & 0.949 \\
CATNet~\cite{Sun2023CATNet}    & 262.73 & 0.932 & 0.937 & 0.960 & 0.938 & 0.934 & 0.971 & 0.910 & 0.928 & 0.951 & 0.920 & 0.922 & 0.958 & 0.952 & 0.958 & 0.975 \\
VST-S++~\cite{Liu2024VSTpp}   & 143.15 & 0.928 & 0.928 & 0.957 & 0.935 & 0.925 & 0.964 & 0.904 & 0.918 & 0.946 & 0.921 & 0.916 & 0.954 & 0.945 & 0.950 & 0.969 \\
CPNet~\cite{Hu2024CMF}     & 216.50 & 0.935 & 0.941 & 0.963 & 0.940 & \textcolor{green}{0.936} & 0.971 & 0.907 & 0.927 & 0.946 & 0.920 & 0.922 & \textcolor{green}{0.960} & 0.951 & 0.959 & 0.974 \\
VSCode-S~\cite{Luo2024VSCode}  & 74.72  & \textcolor{green}{0.944} & \textcolor{green}{0.949} & \textcolor{green}{0.970} & \textcolor{green}{0.941} & 0.932 & 0.968 & \textcolor{green}{0.924} & \textcolor{green}{0.942} & 0.958 & \textcolor{green}{0.931} & \textcolor{green}{0.928} & 0.958 & \textcolor{red}{0.960} & \textcolor{red}{0.967} & \textcolor{red}{0.980} \\
Samba~\cite{He2025Samba} & 54.94 & \textcolor{blue}{0.949} & \textcolor{blue}{0.956} & \textcolor{blue}{0.975} & \textcolor{blue}{0.947} & \textcolor{blue}{0.941} & \textcolor{blue}{0.976} & \textcolor{blue}{0.931} & \textcolor{blue}{0.949} & \textcolor{blue}{0.966} & \textcolor{blue}{0.935} & \textcolor{blue}{0.933} & \textcolor{blue}{0.963} & \textcolor{green}{0.956} & \textcolor{blue}{0.964} & \textcolor{green}{0.976} \\
\midrule
\textbf{DGSSM (Ours)} & \textbf{44.73} & \textbf{\textcolor{red}{0.958}} & \textbf{\textcolor{red}{0.964}} & \textbf{\textcolor{red}{0.982}} & \textbf{\textcolor{red}{0.952}} & \textbf{\textcolor{red}{0.950}} & \textbf{\textcolor{red}{0.983}} & \textbf{\textcolor{red}{0.936}} & \textbf{\textcolor{red}{0.959}} & \textbf{\textcolor{red}{0.971}} & \textbf{\textcolor{red}{0.938}} & \textbf{\textcolor{red}{0.937}} & \textbf{\textcolor{red}{0.968}} & \textcolor{blue}{\textbf{0.959}} & \textcolor{red}{\textbf{0.967}} & \textcolor{blue}{\textbf{0.978}} \\
\bottomrule
\end{tabular}
}
\end{table*}

\begin{table*}[t]
\centering
\caption{Comparison of our DGSSM against other SOTA RGB-T SOD methods on three benchmark datasets. The best results are highlighted in \textcolor{red}{red}, \textcolor{blue}{blue}, and \textcolor{green}{green}, respectively.}
\vspace{-10pt}
\label{tab:rgbt}
\resizebox{\textwidth}{!}{
\begin{tabular}{l|c|ccc|ccc|ccc}
\toprule
\multirow{2}{*}{\textbf{Method}} & \multirow{2}{*}{\textbf{Params (M)}} & \multicolumn{3}{c|}{\textbf{VT821}~\cite{Wang2018RGBTBenchmark}} & \multicolumn{3}{c|}{\textbf{VT1000}~\cite{Tu2019RGBTGraph}} & \multicolumn{3}{c}{\textbf{VT5000}~\cite{Tu2022RGBTDataset}} \\
 & & $S_m\uparrow$ & $F_m\uparrow$ & $E_m\uparrow$ & $S_m\uparrow$ & $F_m\uparrow$ & $E_m\uparrow$ & $S_m\uparrow$ & $F_m\uparrow$ & $E_m\uparrow$ \\
\midrule
\multicolumn{11}{c}{\textit{\textbf{CNN-based}}} \\
\midrule

MIDD~\cite{Tu2021MIDecoder}      & 52.43  & 0.871 & 0.847 & 0.916 & 0.916 & 0.904 & 0.956 & 0.868 & 0.834 & 0.919 \\
CSRNet~\cite{Huo2021ECSRNet}    & 87.09  & 0.885 & 0.855 & 0.920 & 0.919 & 0.901 & 0.952 & 0.868 & 0.821 & 0.912 \\
TNet~\cite{Cong2022Thermal}       & -      & 0.899 & 0.885 & 0.936 & 0.929 & 0.921 & 0.965 & 0.895 & 0.864 & 0.936 \\
CGMDR~\cite{Chen2022CGMDRNet}      & -      & 0.894 & 0.872 & 0.932 & 0.931 & 0.927 & 0.966 & 0.896 & 0.877 & 0.939 \\
SPNet~\cite{Zhang2023SaliencyPrototype}     & 104.03 & 0.913 & 0.900 & 0.949 & 0.941 & 0.943 & 0.975 & 0.914 & \textcolor{green}{0.905} & 0.954 \\
\midrule
\multicolumn{11}{c}{\textit{\textbf{Transformer-based}}} \\
\midrule

SwinNet-B~\cite{Liu2021SwinNet} & 199.18 & 0.904 & 0.877 & 0.937 & 0.938 & 0.933 & 0.974 & 0.912 & 0.885 & 0.944 \\
VSCode-S~\cite{Luo2024VSCode}  & 74.72  & \textcolor{green}{0.926} & \textcolor{green}{0.910} & \textcolor{green}{0.954} & \textcolor{green}{0.952} & \textcolor{green}{0.947} & \textcolor{green}{0.981} & \textcolor{green}{0.925} & 0.900 & \textcolor{green}{0.959} \\
\textbf{Samba~\cite{He2025Samba}} & 54.94 & \textcolor{blue}{0.934} & \textcolor{blue}{0.927} & \textcolor{blue}{0.965} & \textcolor{blue}{0.953} & \textcolor{blue}{0.956} & \textcolor{blue}{0.983} & \textcolor{blue}{0.928} & \textcolor{blue}{0.919} & \textcolor{blue}{0.963} \\
\midrule
\textbf{DGSSM (Ours)} & \textbf{44.73} & \textbf{\textcolor{red}{0.941}} & \textbf{\textcolor{red}{0.939}} & \textbf{\textcolor{red}{0.971}} & \textbf{\textcolor{red}{0.956}} & \textbf{\textcolor{red}{0.966}} & \textbf{\textcolor{red}{0.987}} & \textbf{\textcolor{red}{0.932}} & \textbf{\textcolor{red}{0.922}} & \textbf{\textcolor{red}{0.965}} \\
\bottomrule
\end{tabular}
\vspace{-15pt}

}
\end{table*}
\section{Experiments}
\vspace{-8pt}
We conduct comprehensive experiments to evaluate the effectiveness, robustness, 
and generalization ability of the proposed \dgssm\ across unimodal and multimodal 
salient object detection scenarios. In particular, we aim to answer three questions: 
(i) whether \dgssm\ consistently outperforms existing state-of-the-art methods across 
diverse datasets, (ii) how it compares with recent CNN-, transformer-, and 
state space–based approaches under comparable model complexity, and (iii) whether 
the proposed diffusion-guided formulation scales effectively to RGB-D and RGB-T 
settings. 
\vspace{-8pt}
\subsection{Experimental Setup}
\vspace{-2pt}
\subsection{Datasets.}
\vspace{-5pt}
For RGB salient object detection, we evaluate on five widely used benchmarks: 
DUTS~\cite{Wang2017ImageLevel}, DUT-OMRON~\cite{Yang2013GraphManifold}, HKU-IS~\cite{Li2015MultiScale}, PASCAL-S~\cite{Li2014Secrets}, and ECSSD~\cite{Yan2013Hierarchical}. RGB-D experiments are conducted on 
NJUD~\cite{Ju2014DepthSaliency}, NLPR~\cite{Peng2014RGBD}, SIP~\cite{Fan2020RGBDReview}, STERE~\cite{Niu2012Stereopsis}, and DUTLF-D~\cite{Piao2019DMSRAN}, while RGB-T evaluations are performed on 
VT821~\cite{Wang2018RGBTBenchmark}, VT1000~\cite{Tu2019RGBTGraph}, and VT5000~\cite{Tu2022RGBTDataset}. All experiments follow the standard training and testing 
splits provided by the original datasets.
\vspace{-15pt}

\subsection{Metrics.}
\vspace{-7pt}
Following common practice, we report the structure measure ($S_m$), mean 
F-measure ($F_m$), and mean E-measure ($E_m$). Higher values indicate better 
performance across all metrics.
\vspace{-9pt}
\subsection{Baselines.}
\vspace{-7pt}
We compare \dgssm\ with representative CNN-based, transformer-based, and recent 
state space–based methods. All reported results are taken from the original 
papers or reproduced using publicly available implementations.
\vspace{-9pt}
\subsection{Comparison on RGB Salient Object Detection}
\vspace{-7pt}
Table~\ref{tab:rgb} reports quantitative comparisons on RGB benchmarks. 
\dgssm\ consistently achieves the best performance across all five datasets, 
surpassing strong CNN-based and Transformer-based baselines by a clear margin. 
Notably, \dgssm\ improves $S_m$ and $E_m$ on DUTS~\cite{Wang2017ImageLevel} and ECSSD~\cite{Yan2013Hierarchical}, indicating superior 
structural consistency and boundary preservation in complex scenes.

Compared with recent Transformer-based models such as VSCode-S and Samba, 
\dgssm\ delivers further gains while employing fewer parameters. This improvement 
can be attributed to the diffusion-guided structural prior and iterative refinement, 
which enable more accurate saliency localization without relying on heavy 
self-attention mechanisms. The results on DUT-OMRON~\cite{Yang2013GraphManifold} and PASCAL-S~\cite{Li2014Secrets} further demonstrate 
that \dgssm\ remains robust under challenging background clutter and low-contrast 
conditions.

\subsection{Comparison on RGB-D Salient Object Detection}

Quantitative results on RGB-D benchmarks are summarized in Table~\ref{tab:rgbd}. 
\dgssm\ consistently outperforms all competing methods across NJUD~\cite{Ju2014DepthSaliency}, NLPR~\cite{Peng2014RGBD}, SIP~\cite{Fan2020RGBDReview}, 
STERE~\cite{Niu2012Stereopsis}, and DUTLF-D~\cite{Piao2019DMSRAN} in terms of $S_m$, $F_m$, and $E_m$. In particular, the gains on 
NJUD~\cite{Ju2014DepthSaliency} and NLPR~\cite{Peng2014RGBD} highlight the effectiveness of diffusion-guided priors in stabilizing 
saliency estimation when depth cues are noisy or incomplete.

Compared with recent Transformer-based approaches such as VSCode-S and Samba, 
\dgssm\ achieves superior performance with a more compact parameter budget. 
These results indicate that explicitly modeling deterministic state space dynamics 
together with stochastic diffusion guidance provides a favorable balance between 
accuracy and efficiency for RGB-D saliency detection.
\vspace{-11pt}
\subsection{Comparison on RGB-T Salient Object Detection}
\vspace{-4pt}
Table~\ref{tab:rgbt} presents quantitative comparisons on RGB-T benchmarks. 
\dgssm\ achieves the best results on all three datasets, with particularly strong 
performance on VT1000~\cite{Tu2019RGBTGraph} and VT5000~\cite{Tu2022RGBTDataset}. These datasets contain numerous low-illumination 
and thermal-dominant scenes, where appearance cues alone are insufficient.

The consistent improvements over both CNN-based and Transformer-based methods 
suggest that the proposed framework effectively leverages complementary thermal 
information. Moreover, compared with the recent Samba model, \dgssm\ yields further 
gains in all evaluation metrics, demonstrating that diffusion-guided refinement 
provides additional benefits beyond pure state space modeling.
\vspace{-11pt}
\subsection{Model Complexity Analysis}
\vspace{-4pt}
Despite incorporating diffusion guidance and iterative refinement, \dgssm\ remains 
parameter-efficient. As shown in Tables~\ref{tab:rgb}–\ref{tab:rgbt}, \dgssm\ employs 
fewer parameters than most Transformer-based competitors while consistently achieving 
higher accuracy. This efficiency stems from operating diffusion processes in a compact 
latent space and from the linear complexity of state space scanning.

\vspace{-11pt}
\subsection{Ablation Study}

We conduct detailed ablation experiments to analyze the contribution of each 
component in \dgssm. All ablations are performed on the DUTS~\cite{Wang2017ImageLevel} benchmark under 
identical training and evaluation settings. Starting from a vanilla 
state space baseline, we progressively introduce diffusion guidance, adaptive 
prompting, selective scanning, boundary refinement, and iterative denoising.

Table~\ref{tab:ablation} reports quantitative results. Each component yields a 
consistent improvement, validating the design choices of the proposed framework.

\begin{table}[t]
\centering
\caption{Ablation study on DUTS~\cite{Wang2017ImageLevel}. Each component contributes positively to the 
overall performance.}
\begin{tabular}{lccc}
\toprule
Configuration & $S_m\uparrow$ & $F_m\uparrow$ & $E_m\uparrow$ \\
\midrule
State-Space Baseline              & 0.923 & 0.914 & 0.952 \\
+ Diffusion Structural Prior (DSP) & 0.931 & 0.921 & 0.961 \\
+ Adaptive Saliency Prompting (ASP)& 0.936 & 0.926 & 0.965 \\
+ Multi-Scale Selective Scan (MS-SS) & 0.940 & 0.931 & 0.969 \\
+ Boundary-Aware Refinement (BARH) & 0.944 & 0.935 & 0.973 \\
+ Iterative Diffusion Refinement (IMDR) & 0.946 & 0.938 & 0.976 \\
\midrule
\textbf{Full DGSSM}                & \textbf{0.947} & \textbf{0.939} & \textbf{0.978} \\
\bottomrule
\end{tabular}
\label{tab:ablation}
\end{table}

Introducing the diffusion structural prior leads to a noticeable gain, 
indicating that diffusion-derived global cues effectively regularize early 
feature representations. Adaptive saliency prompting further improves 
discrimination by dynamically modulating state space transitions. The 
multi-scale selective scan enhances spatial continuity, particularly for large 
or elongated objects. Boundary-aware refinement yields additional improvements 
in $E_m$, confirming its effectiveness in sharpening object contours. Finally, 
iterative Mamba diffusion refinement provides the most consistent boost across 
all metrics, demonstrating the benefit of progressive denoising in saliency 
prediction.

\begin{figure}
    \centering
    \includegraphics[width=1.0\linewidth]{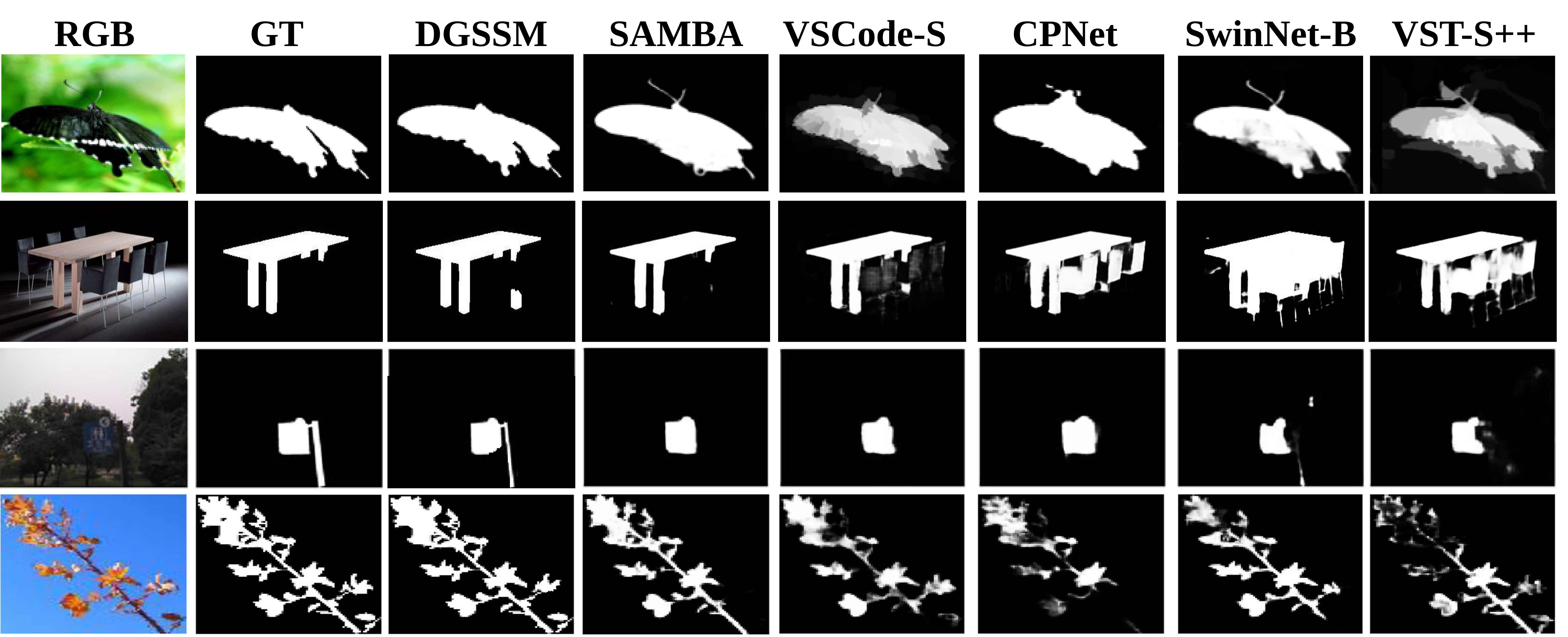}
    \caption{Qualitative Comparison of our DGSSM against SOTA methods}
    \label{fig:qualitative}
\end{figure}


\subsection{Qualitative Analysis}

Figure~\ref{fig:qualitative} presents qualitative comparisons between \dgssm\ 
and representative state-of-the-art methods across RGB, RGB-D, and RGB-T 
scenarios. Competing methods often exhibit fragmented predictions, blurred 
boundaries, or incomplete object regions when confronted with complex 
backgrounds, low contrast, or modality-specific noise.
In contrast, \dgssm\ produces more coherent saliency maps with sharper object 
boundaries and improved completeness. 
The diffusion-guided structural prior 
helps preserve global object layout, while the iterative refinement stage 
effectively suppresses spurious activations and recovers thin structures that 
are frequently missed by other approaches. These advantages are particularly 
pronounced in scenes containing multiple salient objects or severe appearance 
ambiguities.

\section{Conclusion}
This paper introduced \dgssm, a diffusion-guided state space framework for salient 
object detection that unifies deterministic sequential modeling with stochastic 
structural refinement. By reformulating saliency estimation as a progressive denoising 
process, the proposed approach effectively bridges the complementary strengths of 
state space models and diffusion-based priors. The integration of diffusion structural 
guidance, adaptive saliency prompting, multi-scale selective scanning, and iterative 
Mamba diffusion refinement enables \dgssm\ to capture long-range contextual 
dependencies while preserving fine-grained boundary details.
Evaluations on 13 public benchmarks covering RGB, RGB-D, and RGB-T settings 
demonstrate that \dgssm\ consistently outperforms existing state-of-the-art methods 
across multiple metrics, while maintaining a compact parameter footprint. Ablation 
studies further confirm that each component contributes meaningfully to the overall 
performance, and qualitative results highlight improved object completeness and 
boundary sharpness in challenging scenarios involving cluttered backgrounds, low 
contrast, and modality-specific noise.
Beyond salient object detection, the proposed diffusion-guided state space modeling 
paradigm offers a promising direction for dense visual perception tasks that require 
both global reasoning and precise structural localization. Future work may explore 
extending this framework to video-based saliency, instance-level segmentation, and 
other multimodal perception problems.


\begin{thebibliography}{8}

\bibitem{Li2024Depth}
Li, L., Lu, L., et al.: Transformer-based depth optimization network for RGB-D salient object detection. In: \textit{Proceedings of the International Conference on Pattern Recognition (ICPR 2024)}. Springer (2024).

\bibitem{Liang2024Prompt}
Liang, W., et al.: External prompt features enhanced parameter-efficient fine-tuning for salient object detection. In: \textit{Proceedings of the International Conference on Pattern Recognition (ICPR 2024)}. Springer (2024).

\bibitem{Englebert2022BRCAM}
Englebert, A., Cornu, O., De Vleeschouwer, C.: Backward recursive class activation map refinement for high resolution saliency map. In: \textit{Proceedings of the 26th International Conference on Pattern Recognition (ICPR 2022)}. IEEE (2022).

\bibitem{Lin2022Lightweight}
Lin, Y., et al.: A lightweight multi-scale context network for salient object detection in optical remote sensing images. In: \textit{Proceedings of the 26th International Conference on Pattern Recognition (ICPR 2022)}. IEEE (2022).

\bibitem{Zhang2022CSMAN}
Zhang, Y., Hamidouche, W., Deforges, O.: Channel-spatial mutual attention network for 360 salient object detection. In: \textit{Proceedings of the 26th International Conference on Pattern Recognition (ICPR 2022)}. IEEE (2022).


\bibitem{Zhao2019EGNet}
Zhao, J.-X., Liu, J.-J., Fan, D.-P., Cao, Y., Yang, J., Cheng, M.-M.: EGNet: Edge guidance network for salient object detection. In: Proceedings of the IEEE/CVF International Conference on Computer Vision (ICCV), pp. 8779--8788 (2019)

\bibitem{Wu2022EDN}
Wu, Y.-H., Liu, Y., Zhang, L., Cheng, M.-M., Ren, B.: EDN: Salient object detection via extremely-downsampled network. IEEE TIP \textbf{31}, 3125--3136 (2022)

\bibitem{Wang2023PRO}
Wang, Y., Wang, R., Fan, X., Wang, T., He, X.: Pixels, regions, and objects: Multiple enhancement for salient object detection. In: Proceedings of the IEEE/CVF Conference on Computer Vision and Pattern Recognition (CVPR),(2023)

\bibitem{Liu2024VSTpp}
Liu, N., Luo, Z., Zhang, N., Han, J.: VST++: Efficient and stronger visual saliency transformer. IEEE TPAMI (2024)

\bibitem{Ma2023BRF}
Ma, M., Xia, C., Xie, C., Chen, X., Li, J.: Boosting broader receptive fields for salient object detection. IEEE TIP \textbf{32}, 1026--1038 (2023)

\bibitem{Zhuge2022Integrity}
Zhuge, M., Fan, D.-P., Liu, N., Zhang, D., Xu, D., Shao, L.: Salient object detection via integrity learning. IEEE TPAMI \textbf{45}(3), 3738--3752 (2022)

\bibitem{Chen2024RGBD}
Chen, Q., Zhang, Z., Lu, Y., Fu, K., Zhao, Q.: 3-D convolutional neural networks for RGB-D salient object detection and beyond. IEEE TNNLS \textbf{35}(3)(2024)

\bibitem{Fu2020JLDCF}
Fu, K., Fan, D.-P., Ji, G.-P., Zhao, Q.: JL-DCF: Joint learning and densely-cooperative fusion framework for RGB-D salient object detection. CVPR 2020

\bibitem{Hu2024CMF}
Hu, X., Sun, F., Sun, J., Wang, F., Li, H.: Cross-modal fusion and progressive decoding network for RGB-D salient object detection. International Journal of Computer Vision, pp. 1--19 (2024)


\bibitem{Lee2022SPSN}
Lee, M., Park, C., Cho, S., Lee, S.: SPSN: Superpixel prototype sampling network for RGB-D salient object detection. In: Proceedings of ECCV. Springer. (2022)

\bibitem{Sun2023CATNet}
Sun, F., Ren, P., Yin, B., Wang, F., Li, H.: CATNet: A cascaded and aggregated transformer network for RGB-D salient object detection. IEEE Transactions on Multimedia (2023)

\bibitem{Zhou2021Specificity}
Zhou, T., Fu, H., Chen, G., Zhou, Y., Fan, D.-P., Shao, L.: Specificity-preserving RGB-D saliency detection. ICCV, pp. 4681--4691 (2021)

\bibitem{Chen2022CGMDRNet}
Chen, G., Shao, F., Chai, X., Chen, H., Jiang, Q., Meng, X., Ho, Y.-S.: CGMDRNet: Cross-guided modality difference reduction network for RGB-T salient object detection. IEEE TCSVT \textbf{32}(9), 6308--6323 (2022)

\bibitem{Cong2022Thermal}
Cong, R., Zhang, K., Zhang, C., Zheng, F., Zhao, Y., Huang, Q., Kwong, S.: Does thermal really always matter for RGB-T salient object detection? IEEE Transactions on Multimedia \textbf{25}, 6971--6982 (2022)

\bibitem{Huo2021ECSRNet}
Huo, F., Zhu, X., Zhang, L., Liu, Q., Shu, Y.: Efficient context-guided stacked refinement network for RGB-T salient object detection. IEEE TCSVT \textbf{32}(5), 3111--3124 (2021)

\bibitem{Tu2021MIDecoder}
Tu, Z., Li, Z., Li, C., Lang, Y., Tang, J.: Multi-interactive dual-decoder for RGB-thermal salient object detection. IEEE Transactions on Image Processing \textbf{30}, 5678--5691 (2021)


\bibitem{Zhang2023SaliencyPrototype}
Zhang, Z., Wang, J., Han, Y.: Saliency prototype for RGB-D and RGB-T salient object detection. In: Proceedings of the ACM International Conference on Multimedia (ACM MM), pp. 3696--3705 (2023)



\bibitem{Guo2024UNITR}
Guo, R., Ying, X., Qi, Y., Qu, L.: UniTR: A unified transformer-based framework for co-object and multi-modal saliency detection. IEEE Transactions on Multimedia (2024)

\bibitem{Zhao2024MotionMemory}
Zhao, X., Liang, H., Li, P., Sun, G., Zhao, D., Liang, R., He, X.: Motion-aware memory network for fast video salient object detection. IEEE Transactions on Image Processing (2024)

\bibitem{Ji2021FullDuplex}
Ji, G.-P., Fu, K., Wu, Z., Fan, D.-P., Shen, J., Shao, L.: Full-duplex strategy for video object segmentation. ICCV, pp. 4922--4933 (2021)


\bibitem{Liu2023CSTTransformer}
Liu, N., Nan, K., Zhao, W., Yao, X., Han, J.: Learning complementary spatial–temporal transformer for video salient object detection. IEEE Transactions on Neural Networks and Learning Systems (2023)

\bibitem{Lu2022DepthCoop}
Lu, Y., Min, D., Fu, K., Zhao, Q.: Depth-cooperated trimodal network for video salient object detection. In: 2022 IEEE International Conference on Image Processing (ICIP), pp. 116--120. IEEE (2022)

\bibitem{Lin2024VIDSOD100}
Lin, J., Zhu, L., Shen, J., Fu, H., Zhang, Q., Wang, L.: VIDSOD-100: A new dataset and a baseline model for RGB-D video salient object detection. International Journal of Computer Vision, pp. 1--19 (2024)

\bibitem{Mou2024RGBDVideo}
Mou, A., Lu, Y., He, J., Min, D., Fu, K., Zhao, Q.: Salient object detection in RGB-D videos. IEEE Transactions on Image Processing \textbf{33}, 6660--6675 (2024)

\bibitem{Zhu2024VisionMamba}
Zhu, L., Liao, B., Zhang, Q., Wang, X., Liu, W., Wang, X.: Vision Mamba: Efficient visual representation learning with bidirectional state space model. In: Proceedings of the 41st International Conference on Machine Learning (ICML), vol. 235, pp. 62429--62442. PMLR (2024)

\bibitem{Liu2024VMamba}
Liu, Y., Tian, Y., Zhao, Y., Yu, H., Xie, L., Wang, Y., Ye, Q., Liu, Y.: VMamba: Visual state space model. arXiv preprint arXiv:2401.10166 (2024)

\bibitem{Wan2024SIGMA}
Wan, Z., Wang, Y., Yong, S., Zhang, P., Stepputtis, S., Sycara, K., Xie, Y.: SIGMA: Siamese Mamba network for multi-modal semantic segmentation. arXiv preprint arXiv:2404.04256 (2024)

\bibitem{Dong2025FusionMamba}
Dong, W., et al.: Fusion-Mamba for cross-modality object detection. IEEE Transactions on Multimedia (2025)

\bibitem{Zhou2024DMM}
Zhou, M., Li, T., Qiao, C., Xie, D., Wang, G., Ruan, N., Mei, L., Yang, Y.: DMM: Disparity-guided multispectral Mamba for oriented object detection in remote sensing. arXiv preprint arXiv:2407.08132 (2024)

\bibitem{Zhao2024RSMamba}
Zhao, S., Chen, H., Zhang, X., Xiao, P., Bai, L., Ouyang, W.: RS-Mamba for large remote sensing image dense prediction. arXiv preprint arXiv:2404.02668 (2024)

\bibitem{Yang2024PlainMamba}
Yang, C., Chen, Z., Espinosa, M., Ericsson, L., Wang, Z., Liu, J., Crowley, E.J.: PlainMamba: Improving non-hierarchical Mamba in visual recognition. BMVC. (2024)


\bibitem{Mei2024CoDi}
Mei, K., Delbracio, M., Talebi, H., Tu, Z., Patel, V.M., Milanfar, P.: CoDi: Conditional diffusion distillation for higher-fidelity and faster image generation. In: Proceedings of the IEEE/CVF CVPR, pp. 9048--9058 (2024)

\bibitem{Zhang2023ConditionalControl}
Zhang, L., Rao, A., Agrawala, M.: Adding conditional control to text-to-image diffusion models. In: Proceedings of the IEEE/CVF ICCV, pp. 3836--3847 (2023)


\bibitem{Moser2024DiffusionSurvey}
Moser, B.B., Shanbhag, A.S., Raue, F., Frolov, S., Palacio, S., Dengel, A.: Diffusion models, image super-resolution, and everything: A survey. IEEE Transactions on Neural Networks and Learning Systems, pp. 1--21 (2024)


\bibitem{Gao2020Efficient}
S.-H. Gao, Y.-Q. Tan, M.-M. Cheng, C. Lu, Y. Chen, and S. Yan.
\newblock Highly efficient salient object detection with 100k parameters.
\newblock In \emph{Proceedings of the European Conference on Computer Vision (ECCV)},
pages 702--721. Springer, 2020.


\bibitem{Wang2023Multiple}
Y. Wang, R. Wang, X. Fan, T. Wang, and X. He.
\newblock Pixels, regions, and objects: Multiple enhancement for salient object detection.
\newblock In CVPR, 2023.

\bibitem{Luo2024VSCode}
Z. Luo, N. Liu, W. Zhao, X. Yang, D. Zhang, D.-P. Fan, F. Khan, and J. Han.
\newblock VSCode: General visual salient and camouflaged object detection with 2D prompt learning.
\newblock In \emph{Proceedings of the IEEE/CVF CVPR}, pages 17169--17180, 2024.

\bibitem{He2025Samba}
J. He, K. Fu, X. Liu, and Q. Zhao.
\newblock Samba: A unified Mamba-based framework for general salient object detection.
\newblock In \emph{Proceedings of the IEEE/CVF Conference on Computer Vision and Pattern Recognition (CVPR)},
pages 25314--25324, 2025.

\bibitem{Liu2021SwinNet}
Z. Liu, Y. Tan, Q. He, and Y. Xiao.
\newblock SwinNet: Swin transformer drives edge-aware RGB-D and RGB-T salient object detection.
\newblock \emph{IEEE Transactions on Circuits and Systems for Video Technology}, 32(7):4486--4497, 2021.

\bibitem{Song2022MGAN}
K. Song, L. Huang, A. Gong, and Y. Yan.
\newblock Multiple graph affinity interactive network and a variable illumination dataset for RGB-T image salient object detection.
\newblock \emph{IEEE Transactions on Circuits and Systems for Video Technology}, 33(7), 2022.

\bibitem{Wang2017ImageLevel}
L. Wang, H. Lu, Y. Wang, M. Feng, D. Wang, B. Yin, and X. Ruan.
\newblock Learning to detect salient objects with image-level supervision.
\newblock In \emph{Proceedings of the IEEE Conference on Computer Vision and Pattern Recognition (CVPR)}, 2017.

\bibitem{Yang2013GraphManifold}
C. Yang, L. Zhang, H. Lu, X. Ruan, and M.-H. Yang.
\newblock Saliency detection via graph-based manifold ranking.
\newblock In \emph{Proceedings of the IEEE Conference on Computer Vision and Pattern Recognition (CVPR)},
pages 3166--3173, 2013.

\bibitem{Li2015MultiScale}
G. Li and Y. Yu.
\newblock Visual saliency based on multi-scale deep features.
\newblock In \emph{Proceedings of the IEEE Conference on Computer Vision and Pattern Recognition (CVPR)},
pages 5455--5463, 2015.

\bibitem{Li2014Secrets}
Y. Li, X. Hou, C. Koch, J. M. Rehg, and A. L. Yuille.
\newblock The secrets of salient object segmentation.
\newblock In \emph{Proceedings of the IEEE Conference on Computer Vision and Pattern Recognition (CVPR)},
pages 280--287, 2014.

\bibitem{Yan2013Hierarchical}
Q. Yan, L. Xu, J. Shi, and J. Jia.
\newblock Hierarchical saliency detection.
\newblock In \emph{Proceedings of the IEEE Conference on Computer Vision and Pattern Recognition (CVPR)}, 2013.

\bibitem{Ju2014DepthSaliency}
R. Ju, L. Ge, W. Geng, T. Ren, and G. Wu.
\newblock Depth saliency based on anisotropic center-surround difference.
\newblock In \emph{Proceedings of the IEEE International Conference on Image Processing (ICIP)}, pages 1115--1119. IEEE, 2014.

\bibitem{Peng2014RGBD}
H. Peng, B. Li, W. Xiong, W. Hu, and R. Ji.
\newblock RGB-D salient object detection: A benchmark and algorithms.
\newblock In \emph{Proceedings of the European Conference on Computer Vision (ECCV)},
pages 92--109. Springer, 2014.

\bibitem{Fan2020RGBDReview}
D.-P. Fan, Z. Lin, Z. Zhang, M. Zhu, and M.-M. Cheng.
\newblock Rethinking RGB-D salient object detection: Models, data sets, and large-scale benchmarks.
\newblock \emph{IEEE Transactions on Neural Networks and Learning Systems}, 32(5):2075--2089, 2020.

\bibitem{Niu2012Stereopsis}
Y. Niu, Y. Geng, X. Li, and F. Liu.
\newblock Leveraging stereopsis for saliency analysis.
\newblock In \emph{Proceedings of the IEEE Conference on Computer Vision and Pattern Recognition (CVPR)},
pages 454--461. IEEE, 2012.

\bibitem{Piao2019DMSRAN}
Y. Piao, W. Ji, J. Li, M. Zhang, and H. Lu.
\newblock Depth-induced multi-scale recurrent attention network for saliency detection.
\newblock In \emph{Proceedings of the IEEE International Conference on Computer Vision (ICCV)},
pages 7254--7263, 2019.

\bibitem{Wang2018RGBTBenchmark}
G. Wang, C. Li, Y. Ma, A. Zheng, J. Tang, and B. Luo.
\newblock RGB-T saliency detection benchmark: Dataset, baselines, analysis and a novel approach.
\newblock In \emph{Proceedings of the International Conference on Intelligent Graphics and Interactive Techniques (IGTA)},
pages 359--369. Springer, 2018.

\bibitem{Tu2019RGBTGraph}
Z. Tu, T. Xia, C. Li, X. Wang, Y. Ma, and J. Tang.
\newblock RGB-T image saliency detection via collaborative graph learning.
\newblock \emph{IEEE Transactions on Multimedia}, 22(1), 2019.

\bibitem{Tu2022RGBTDataset}
Z. Tu, Y. Ma, Z. Li, C. Li, J. Xu, and Y. Liu.
\newblock RGB-T salient object detection: A large-scale dataset and benchmark.
\newblock \emph{IEEE Transactions on Multimedia}, 2022.

\bibitem{Zhou2021SaliencyTracking}
Z. Zhou, W. Pei, X. Li, H. Wang, F. Zheng, and Z. He.
\newblock Saliency-associated object tracking.
\newblock In \emph{Proceedings of the IEEE/CVF ICCV}, 2021.

\bibitem{Miangoleh2023SaliencyEnhancement}
S. M. H. Miangoleh, Z. Bylinskii, E. Kee, E. Shechtman, and Y. Aksoy.
\newblock Realistic saliency guided image enhancement.
\newblock In \emph{Proceedings of the IEEE/CVF Conference on Computer Vision and Pattern Recognition (CVPR)},
pages 186--194, 2023.

\bibitem{Jiang2024LFTransformer}
Y. Jiang, X. Li, K. Fu, and Q. Zhao.
\newblock Transformer-based light field salient object detection and its application to autofocus.
\newblock \emph{IEEE Transactions on Image Processing}, 33:6647--6659, 2024.

\bibitem{Jiang2024VLMAssessment}
Y. Jiang, X. Yan, G.-P. Ji, K. Fu, M. Sun, H. Xiong, D.-P. Fan, and F. S. Khan.
\newblock Effectiveness assessment of recent large vision-language models.
\newblock \emph{Visual Intelligence}, 2(1):17, 2024.


\bibitem{Liu2021Swin}
Z. Liu, Y. Lin, Y. Cao, H. Hu, Y. Wei, Z. Zhang, S. Lin, and B. Guo.
\newblock Swin Transformer: Hierarchical vision transformer using shifted windows.
\newblock In \emph{Proceedings of the IEEE/CVF International Conference on Computer Vision (ICCV)},
pages 10012--10022, 2021.

\bibitem{Mehta2022MobileViT}
S. Mehta and M. Rastegari.
\newblock MobileViT: Lightweight, general-purpose, and mobile-friendly vision transformer.
\newblock In \emph{Proceedings of the International Conference on Learning Representations (ICLR)},
2022.

\bibitem{Gu2023Mamba}
A. Gu and T. Dao.
\newblock Mamba: Linear-time sequence modeling with selective state spaces.
\newblock \emph{arXiv preprint arXiv:2312.00752}, 2023.






\end{thebibliography}
\end{document}